\documentclass[letterpaper, 10 pt, conference]{ieeetj}  

\usepackage{cite}
\usepackage{amsmath,amssymb,amsfonts}
\usepackage{algorithmic}
\usepackage{graphicx,color}
\usepackage{textcomp}
\usepackage{xcolor}
\usepackage{hyperref}
\usepackage{algorithm,algorithmic}
\def\BibTeX{{\rm B\kern-.05em{\sc i\kern-.025em b}\kern-.08em
    T\kern-.1667em\lower.7ex\hbox{E}\kern-.125emX}}
\AtBeginDocument{\definecolor{tmlcncolor}{cmyk}{0.93,0.59,0.15,0.02}\definecolor{NavyBlue}{RGB}{0,86,125}}

\usepackage{float}
\usepackage{graphics} 
\usepackage{epsfig} 
\usepackage{tikz}
\usetikzlibrary{calc}
\usetikzlibrary{arrows}
\usepackage{siunitx}

\usepackage{todonotes}
\usepackage{capt-of} 
\usepackage{booktabs}
\newcommand{\argmin}{\operatornamewithlimits{arg\ min}}



\def\OJlogo{\vspace{-4pt}$<$Society logo(s) and publication title will appear here.$>$}
\def\seclogo{\vspace{10pt}$<$Society logo(s) and publication title will appear here.$>$}

\def\authorrefmark#1{\ensuremath{^{\textbf{#1}}}}

\begin{document}
\receiveddate{XX Month, XXXX}
\reviseddate{XX Month, XXXX}
\accepteddate{XX Month, XXXX}
\publisheddate{XX Month, XXXX}
\currentdate{XX Month, XXXX}
\doiinfo{XXXX.2022.1234567}

\markboth{}{Author {et al.}}

\title{Sapling-NeRF: Geo-Localised Sapling Reconstruction in Forests for Ecological Monitoring}


\author{Miguel Ángel Muñoz-Bañón\authorrefmark{1.2},  Nived Chebrolu\authorrefmark{1},  Sruthi M. Krishna Moorthy\authorrefmark{3}, Yifu~Tao\authorrefmark{1}, Fernando Torres\authorrefmark{2}, Roberto Salguero-Gómez\authorrefmark{3} and Maurice Fallon\authorrefmark{1}}

\affil{Oxford Robotics Institute, Department of Engineering Science, University of Oxford, Oxford, UK}
\affil{Group of Automation, Robotics and Computer Vision, University of Alicante, Alicante, Spain}
\affil{Department of Biology, University of Oxford, Oxford, UK}

\corresp{Corresponding author: Miguel Ángel Muñoz-Bañón (email: miguelangel.munoz@ua.es).}

\authornote{This work is supported in part by the EU Horizon 2020 Project 101070405 (DigiForest) and a Royal Society University \\Research Fellowship. Miguel Ángel Muñoz-Bañón is supported by the Valencian Community Government and the \\European Union through the fellowship~CIAPOS/2023/101}

\begin{abstract}
Saplings are key indicators of forest regeneration and overall forest health. However, their fine-scale architectural traits are difficult to capture with existing 3D sensing methods, which make quantitative evaluation difficult. Terrestrial Laser Scanners (TLS), Mobile Laser Scanners (MLS), or traditional photogrammetry approaches poorly reconstruct thin branches, dense foliage, and lack the scale consistency needed for long-term monitoring. Implicit 3D reconstruction methods such as Neural Radiance Fields (NeRF) and 3D Gaussian Splatting (3DGS) are promising alternatives, but cannot recover the true scale of a scene and lack any means to be accurately geo-localised. In this paper, we present a pipeline which fuses NeRF, LiDAR SLAM, and GNSS to enable repeatable, geo-localised ecological monitoring of saplings. Our system proposes a three-level representation: (i) coarse Earth-frame localisation using GNSS, (ii) LiDAR-based SLAM for centimetre-accurate localisation and reconstruction, and (iii) NeRF-derived object-centric dense reconstruction of individual saplings. This approach enables repeatable quantitative evaluation and long-term monitoring of sapling traits. Our experiments in forest plots in Wytham Woods (Oxford, UK) and Evo (Finland) show that stem height, branching patterns, and leaf-to-wood ratios can be captured with increased accuracy as compared to TLS. We demonstrate that accurate stem skeletons and leaf distributions can be measured for saplings with heights between \SI{0.5}{\meter} and \SI{2}{\meter} \emph{in situ}, giving ecologists access to richer structural and quantitative data for analysing forest dynamics.
\end{abstract}



\maketitle

\section{INTRODUCTION}
\label{sec:introduction}

Saplings play a crucial ecological role in forest dynamics, serving as key indicators of forest growth potential. They form the recruitment pool from which mature, canopy-dominating trees eventually emerge. These trees define the future composition, structure, and diversity of a forest. The architectural traits of saplings include stem height, branching patterns, leaf distribution, and leaf-to-wood ratios. They are key indicators of the immediate survival and growth of a sapling: they dictate a sapling's efficiency in capturing light, transporting water, maintaining mechanical stability, and optimising growth strategies in diverse light environments~\cite{poorter1999light}. In addition, these traits also influence broader ecological processes such as competition for resources, habitat creation, and biodiversity maintenance~\cite{kohyama1990significance}. These architectural attributes vary significantly between species, and are shaped by adaptive strategies to specific ecological niches, whether in shaded understories or sunlit canopy gaps.

\begin{figure*}[t]
    \centering
    \includegraphics[]{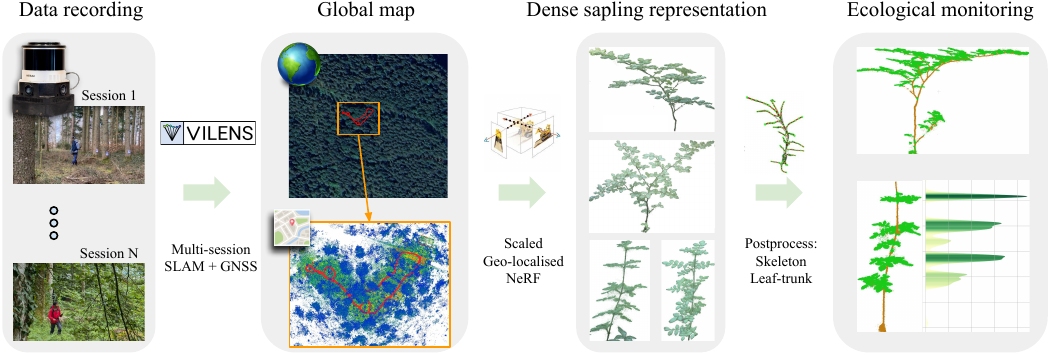}
    \caption{\textbf{We propose a joint NeRF and LiDAR SLAM system for sapling reconstruction \emph{in situ} in the forest using data from a handheld sensing device}. 1) The core LiDAR SLAM system supports hectare-scale multi-session map merging as well as GNSS alignment. 2) We extract a sub-trajectory from a mapping session which encircles an individual sapling of interest and run Structure-from-Motion (SfM, COLMAP) while using the SLAM-derived trajectory to determine consistent metric scale and global localisation. The images and poses are then used to train NeRF models for each sapling. 3) Finally, we demonstrate that a NeRF-derived point cloud s sufficiently accurate to detect the tree skeleton and to measure leaf-wood separation and the leaf distribution. In Section \ref{sec:ev_pc} we demonstrate longitudinal monitoring of saplings - from summer to winter.}
    \label{fig:hero}
\end{figure*}%

Despite the ecological importance of the sapling architecture measurements, precisely capturing these fine-scale structural traits in the field remains a significant challenge. Established methods using Terrestrial Laser Scanning (TLS)~\cite{liang2016terrestrial} are unable to capture fine structural details of saplings, such as thin branches and foliage, due to occlusion and insufficient spatial resolution~\cite{cerbone2025democratizing}. Classic photogrammetry techniques produce sparse reconstruction results, especially within complex, natural habitats. Additionally, these approaches rely on imprecise GNSS fixes to localise the sapling in a common coordinate system - with a new coordinate frame established from scratch for each SfM session. As purely visual mapping systems they also suffer from scale ambiguity. This makes automatic long-term sapling monitoring difficult, especially when months or years pass between data recordings.

Emerging technologies such as Neural Radiance Fields (NeRF)~\cite{wang2024nerf} and 3D Gaussian Splatting (3DGS)~\cite{zhu20243d} are capable of producing dense reconstructions using relatively few views. Prior works have demonstrated reconstruction of fine detail such as bike spokes and cables as well as being able to infer the location of ambient lighting. Thus, NeRF and 3DGS are promising approaches sapling reconstruction. Some works have shown promis in controlled indoor conditions~\cite{adebola2025growsplat,wu20253d}, however it is still challenging to use these technologies \emph{in situ} in natural environments due to light conditions, wind, and the difficulty of recording data around a sapling without actually disturbing the plant and blurring the representation. Furthermore, these vision-based approaches cannot determine the underlying metric scale of the scene, making it impossible to determine quantitative ecological metrics and to track them over time.

To overcome the limitations mentioned above, we propose to loosely couple NeRF reconstruction with LiDAR-based SLAM (Simultaneous Localisation and Mapping). We combining the work with GNSS information (Figure~\ref{fig:hero}), we can then represent the saplings with 3 levels of positioning and reconstruction: imprecise Earth frame location; precise proximal location (and coarse 3D reconstruction) within a LiDAR map of the forest; and detailed NeRF-derived dense reconstruction and novel-view~synthesis.

This strategy has two key advantages: (i) \textbf{Large-scale and detailed representation}: It can build object-centric NeRF-derived reconstruction of specific plants/saplings \footnote{Our technique could also be used to capture animal habitats or watering holes.}. This is in contrast with
other approaches that build scene-centric reconstructions by dividing the environment spatially and creating a submap for each cell~\cite{tancik2022block}. Our approach can build efficient yet precise NeRF models which capture the fine details of the saplings with less attention given to the background. We can embed these NeRF-derived sapling reconstructions within the LiDAR-based reconstruction of a larger hectare-sized plot, which accurately captures the geometry of the background. (ii)~\textbf{Long-term representation}: Our SLAM system can combine multiple sessions recorded over extended time periods (months or years), allowing us to record individual saplings as they grow and evolve. This allows to obtain metrics such as tree structure or leaf distribution to be monitored over multiple seasons or potentially years.

In summary, our contributions are the following:

\begin{itemize}
    \item Object-centric NeRF-derived reconstruction of saplings in challenging natural environments, which allows structural details of the saplings to be captured, which is typically challenging for TLS and MLS-based reconstruction.
    \item A direct linkage between the NeRF output and a multi-session geo-referenced SLAM system, which defines the geometric scale of the NeRFs and their centimetre-accurate localisation which allows monitoring of individual saplings over multiple seasons.
    \item Demonstrations showing that stem height, branching patterns, leaf distribution, and leaf-to-wood ratios can be measured for small saplings (\SI{0.5}{\meter}-\SI{2}{\meter} in height) \emph{in situ} using this approach, achieving improved accuracy compared to those obtained using TLS.
\end{itemize}

The rest of this paper is organised as follows: In Section~\ref{sec:review}, we review related works. In Section \ref{sec:system}, we provide an overview of the entire system, with each module described in Section \ref{sec:methodology}. In Section \ref{sec:evaluation}, we evaluate each module, as well as demonstrating the measurement of different sapling architectural traits.


\section{RELATED WORK}
\label{sec:review}

In recent years, there has been increasing interest in using robotics and computer vision to monitor plant and tree health in forests~\cite{mattamala2025forest}. In Section~\ref{sec:review_plants}, we review recent research on radiance field reconstruction and its application for trees and plant modelling. In Section~\ref{sec:review_scale} we review methods for radiance field reconstruction of large-scale environments as well as their use in long-term monitoring.

\subsection{Radiance Fields for trees and plants}
\label{sec:review_plants}

Techniques for digital forest modelling have predominantly relied on LiDAR technology and focused on reconstructing 3D models of mature trees to measure structural attributes such as height, crown volume, branch topology, and biomass~\cite{borsah2023lidar}. However, capturing saplings accurately remains challenging, because expensive survey-grade LiDAR sensors are typically too sparse to capture fine details such as leaves and small branches (smaller than about \SI{1}{\centi\meter}).

Neural radiance fields (NeRF), introduced in the seminal work by Mildenhall et al.~\cite{mildenhall2021nerf}, demonstrated that using accurately posed images one can reconstruct a dense textured scene representation. This representation can be used to synthesise photorealistic novel-view images capturing fine details of small objects. Subsequent improvements have extended the approach to better handle unbounded scenes~\cite{barron2022mipnerf360}, aliasing~\cite{barron2021mipnerf}, and to have faster training and inference speed using explicit representations such as voxels~\cite{yu2021plenoctrees} and hash grids~\cite{muller2022instant}. An alternative, 3D Gaussian Splatting, has gained popularity since it combines rasterisation and uses an explicit 3D Gaussian representations. 3DGS achieves highly-efficient test-time rendering --- much faster than prior works~\cite{yu2021plenoctrees,muller2022instant}. Follow-up works include surface regularisation~\cite{huang20242dgs} and anti-aliasing features~\cite{yu2024mipsplat} to improve the geometry and rendering quality.

The precise, detailed and visually accurate reconstructions created by radiance field approaches make them appealing for quantitative ecological analysis and effective forest monitoring~\cite{cerbone2025democratizing}. In \cite{korycki2025nerf}, the authors used NeRF-derived reconstructions to measure the Diametre-at-Breast-Height (DBH) of individual trees. However, this approach was not been demonstrated in large-scale scenes or for long-term monitoring. Although DBH can be performed with Mobile Laser Scanners (MLS)~\cite{tremblay2020automatic,tang2015slam}, Korycki's work demonstrates that a simple low-cost camera can measure this parameter. Regardless, we feel that the potential of NeRF to reconstructe fine tree details has not get been sufficiently explored.

Other works~\cite{huang2024evaluating,masiero2024comparing} have compared NeRF-derived reconstruction to LiDAR-based approaches for mature trees, but not the fine structures of saplings or plants. In \cite{adebola2025growsplat}, the authors exploit the capabilities of 3D Gaussian Splatting (3DGS) to monitor the growth of small trees. Different from the aforementioned works which target controlled indoor facilities~\cite{adebola2025growsplat} and parks~\cite{huang2024evaluating,masiero2024comparing}, our work instead targets forest environments which have poor lighting conditions, rough terrain and complicated data recording logistics. Our approach can process measurements taken \emph{in situ} in a forest, allowing the monitoring of saplings within large hectare-scale plots over extended periods.

In this work, we use a NeRF-based approach because we prioritise memory efficiency (to support larger representations) over faster rendering speed. 3D Gaussian Splatting would be an appropriate alternative but requires a larger memory footprint.

\subsection{Large-scale and long-term Radiance Fields}
\label{sec:review_scale}

Many works that extend NeRFs to large-scale environments adopt a submapping approach, where each submap is presented as an individual radiance field model. Block-NeRF~\cite{tancik2022block} proposed to reconstruct a large-scale urban environment by dividing the scene at road intersections and later merging the submaps to achieve continuous image rendering. Mega-NeRF~\cite{meganerf} adopts a grid-based partitioning strategy for large-scale environments. SiLVR~\cite{tao2025silvr} partitions sensor trajectories based on visibility and filters reconstruction artifacts based on uncertainty estimates when merging submaps. Methods that extend the scalability of 3D Gaussian Splatting have used hierarchical structures~\cite{hierarchicalgaussians24} such as Octrees~\cite{ren2024octree}, which enable efficient and high-fidelity rendering at different levels of detail. Different from these works, which focus on reconstructing open environments, our work focuses on regions of interest (the saplings) and localises them within a globally consistent point cloud map built by a LiDAR SLAM system.

Finally some works have sought to capture the dynamics of environments~\cite{pumarola2021d,guo2024motion}. These approaches focus on dynamic objects moving in front of the camera and focus only on short time frames. For longer-term monitoring, recent research has focused on object-level change detectionl~\cite{lu20253dgs,huang2025semanticdifference}. In our work, we implement a different strategy. We have the objects (the sapling) geo-localised, and we are interested in monitoring how their structural traits change over time such as height and leaf distribution.

\begin{figure}[t]
\centering
\includegraphics[]{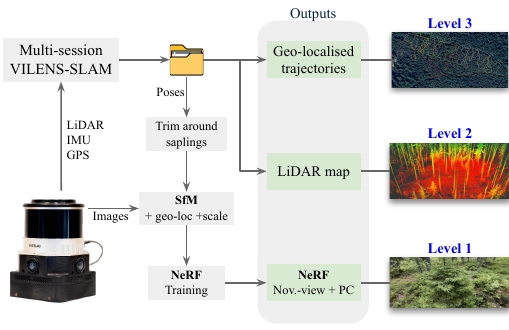}
\caption{The proposed system maps a forest environment at three levels of representation. Levels 3 is coarse GNSS-based localisation. Level 2 is a centimetre-accurate representation from a own multi-session LiDAR SLAM system while Level 1 is generated using images captured around a sapling, using SfM to estimate image poses and localising and scaling those poses according to the SLAM system, and finally training a scale-consistent, geo-localised NeRF model for each sapling.}
\label{fig:system_overview}
\end{figure}

\section{SYSTEM OVERVIEW}
\label{sec:system}

The main idea of this work is to augment lidar maps with NeRF-derived reconstructions of individual saplings to support their long-term monitoring within large hectare-scale plots. To approach this problem, we used the prototype mapping device shown in Figure \ref{fig:system_overview}, which contains LiDAR, IMU, GNSS sensors as well as three cameras. The device can be mounted on a robot, but in this work it was hand-carried. We first build a plot-level forest reconstruction by walking through the forest in a 'lawnmower' pattern, which is suitable for generating plot-level 3D LiDAR maps. We then augment the representation by recording dense camera views cover the entirity of a sapling from all surrounding directions. To achieve this, when we arrive at a sapling that we want to represent with a NeRF, we scan the tree with an inwards-facing 'dome' pattern circumscribing the tree (Figure~\ref{fig:camera-poses}). The data acquisition for this can be challenging in a forest when there is understory and branches close to the saplings. A lack of care during data acquisition can affect the quality of the final reconstruction.

To integrate the dense NeRF-derived reconstruction of the sapling into a consistent geo-localised, multi-session metric map, we loosely couple the NeRF pipeline with a multi-session SLAM pipeline. In Figure~\ref{fig:system_overview}, we show an overview of the proposed system. Conceptually, we divide our representation into three levels of abstraction. \textbf{Level 3} is at Earth level, using 2\,Hz GNSS. \textbf{Level 2} is a point cloud map of a complete forest plot where previous work has used the cloud to extract a forest inventory of individual trees, their diameters, species and heights. Mobile Laser Scanners (MLS) are however not accurate enough to capture fine detail about the structure of small objects such as saplings. This is addressed by \textbf{Level 1} that represents the dense NeRF-derived reconstructions of the saplings which capture detailed point clouds and allow for novel-view synthesis.

\section{METHODOLOGY}
\label{sec:methodology}

In this section, we provide a detailed description of each part of the system. In Section~\ref{sec:slam}, we describe the multi-session SLAM system used for plot-level LiDAR mapping and trajectory estimation (Levels 3 and 2). Section~\ref{sec:nerf} presents the process of obtaining a NeRF-derived representation that is metrically scaled and geo-localised by combining with the SLAM system output (Level 1). Given the dense point clouds obtained from the NeRF, we describe in Section~\ref{sec:skeleton} how existing methods for branch skeletonisation can be used to obtain metrics useful for monitoring the growth and health of saplings.

\begin{figure}[t]
\centering
\includegraphics[]{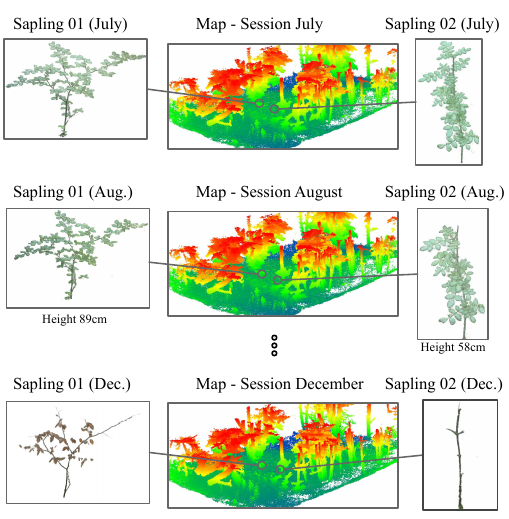}
\caption{By co-registering multiple mapping sessions, we can create a unified map made up of trajectories across different time periods. In this way, we can monitor saplings from summer to winter.}
\label{fig:slam_system}
\end{figure}

\subsection{Multi-session and geo-referenced SLAM}
\label{sec:slam}

The core of the proposed approach is VILENS-SLAM, an online pose graph based LiDAR SLAM system which combines LiDAR-Inertial Odometry~\cite{wisth2022vilens} with a place recognition module~\cite{ramezani2020online} to identify loop closure constraints.

The maps and trajectories from individual SLAM sessions can be merged into a single multi-session pose graph map offline using the approach described in \cite{oh2024evaluation}. It is worth noting that by merging the maps at the pose graph level the point cloud maps are deformed and kept locally consistent. In Figure~\ref{fig:slam_system}, we depict the strategy of using the multi-mission SLAM system to monitor individual saplings. The $i^{\text{th}}$ SLAM session is comprised of a map $\mathcal{M}^{M_i}$, a set of LiDAR scans~$\mathbf{P}^L_i$ and a set of poses:

\begin{equation}
    \mathbf{x}^{M_i}_i = (x_{i_1}, \dots x_{i_k}), \quad i=1,\dots,N, \quad x \in SE(3).
    \label{eq:trajectory}
\end{equation}

The super-index $M_i$ indicates that each session is represented with respect to its own map coordinate frame.

The pose graph map from each session is fed to the multi-session SLAM module, where each individual point cloud $P \in \mathbf{P}^L_i$ and its associated pose $x \in \mathbf{x}^{M_i}_i$ is used by the place recognition module to identify new loop closures. When a successful loop closure constraint is found, it is added to the combined pose graph using the first session as a reference frame. This process generates a single combined point cloud map $\mathcal{M}^{M_1}$ and a combined set of trajectories:

\begin{equation}
    \mathbf{X}^{M_1} = (\mathbf{x}_1, \dots, \mathbf{x}_N).
    \label{eq:trajectories}
\end{equation}

Separately from this process, we also estimate a single transformation between the map frame $M_1$ and the earth frame $E$. This is done in a loosely coupled manner. To estimate the transformation $\mathbf{T}^{E}_M = (\mathbf{R}, \mathbf{t})$, we first associate a GNSS latitude and longitude measurement with each pose. We transform those coordinates into a local Northing and Easting frame in metres. To implement this, we associate the first $u^{\text{th}}$ poses with the first $u^{\text{th}}$ GNSS measurements in $g$ using the timestamps that are closest to each other. We then search for the transformation that minimises the error between the positions in the trajectory and the GNSS observations:

\begin{equation}
    \mathbf{T}^{E*}_M = \argmin_{\mathbf{T}} \left\lVert (\mathbf{R} \mathbf{x}^{x,y}_{1:u} + \mathbf{t}) - \mathbf{g}_{1:u} \right\lVert^2_{2}.
    \label{eq:optimization}
\end{equation}

This allows us to determine the location of each local pose in the global coordinate frame as well as the linked point cloud and semantic details. This information is crucial for long-term inventory and monitoring (of mature trees). We have found our GNSS alignment to be accurate to about one metre, with GNSS reception only affected in the densest forest canopy.

Sections of each trajectory will correspond to where individual saplings have been circumscribed for image capture. We index each sapling within a plot as $j = 1, \dots, M$. Then, for each $\mathbf{x}^{M_1}_i$ trajectory, we can define a set of subtrajectories as:

\begin{equation}
    \mathbf{Y}^{M_1}_i = (\mathbf{y}_{i_1}, \dots, \mathbf{y}_{i_M}).
    \label{eq:subtrajectories}
\end{equation}

In this work, the sapling subtrajectory was extracted manually, but for future work we plan to log the start and end of the sapling during scanning. The methodology described in this section provides the structure to cover saplings in large-scale environments for long-term monitoring. In the next section, we explain how to produce NeRF-derived reconstructions using the images associated with the subtrajectories $ \mathbf{Y}^{M_1}_i$.

\begin{figure}[ht]
\centering
\includegraphics[]{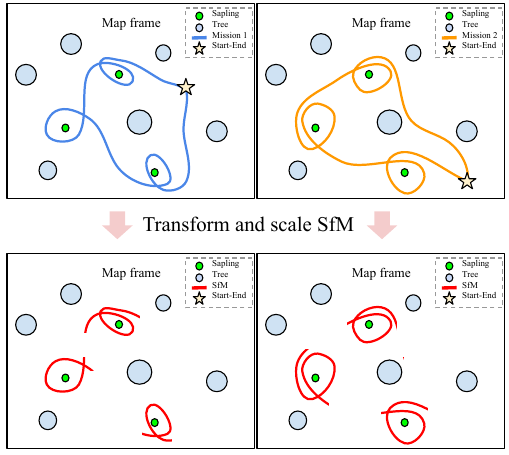}
\caption{\textbf{Rescaling and aligning multiple SfM reconstructions}: From each SLAM-derived session of a plot (upper), we extract a subtrajectory where a sapling is scanned in detail, and then estimate a COLMAP SfM trajectory which has a scale ambiguity. This is corrected using the SLAM-derived trajectory (lower) to produce the (co-aligned) trajectories shown in red in the example. }
\label{fig:sfm_coupled}
\end{figure}

\subsection{NeRF with loose coupling to LiDAR SLAM}
\label{sec:nerf}

To obtain a detailed representation of each sapling, we use Neural Radiance Fields (NeRF)~\cite{wang2024nerf}. NeRF pipelines can render novel views of a scene and also generate dense colorised point clouds. Each sapling is small relative to the size of hectare-scale forest plot. Thus we take the approach of training a single NeRF model for each sapling by scanning it from all directions. The NeRF training pipeline requires a set of images of the scene with precise pose estimates. As we mentioned in the previous section, our SLAM system produces a trajectory around each $j^{\textbf{th}}$ sapling from each $i^{\textbf{th}}$ trajectory and we also have good synchronization between our LiDAR data and our camera image. However, we have found that the poses obtained by the SLAM system are not sufficiently precise to achieve the highest possible quality NeRF-derived reconstruction. To refine the sensor poses, we use Structure from Motion (SfM), which is a widely used approach to estimate the input poses for NeRF. Specifically, we use COLMAP~\cite{schoenberger2016sfm,schoenberger2016mvs} to produce a trajectory around each sapling: 


\begin{equation}
    \mathbf{Z} = (\mathbf{z}^{F_{11}}_{11}, \dots, \mathbf{z}^{F_{NM}}_{NM}).
    \label{eq:colmap_trajectories}
\end{equation}

The superscript $F$ means that each new sapling scan trajectory is represented in it's own local SfM frame. However, because COLMAP is a monocular vision-based approach, this trajectory is only accurate up to a scale ambiguity. To represent each refined trajectory in the map frame $M_1$ and to resolve the correct scaling factor, we use the Umeyama method~\cite{umeyama1991least}. 




By following this process, the pipeline can obtain a set of precise camera poses, represented in a single consistent coordinate frame $M_1$, as depicted in red in Figure~\ref{fig:sfm_coupled}. Then for each $i^{\textbf{th}}$ trajectory, we have:

\begin{equation}
    \mathbf{Z}^{M_1}_i = (\mathbf{z}_{i_1}, \dots, \mathbf{z}_{i_M}).
    \label{eq:scaled_subtrajectories}
\end{equation}

Given $\mathbf{Z}^{M_1}$ trajectories and their associated camera images, we then train a NeRF model for each sapling. In this way, we can generate dense, scaled and geo-referenced point clouds $\mathbf{S}_{ij}$ in the LiDAR map frame and position them with the larger hectare-scale plots.

\subsection{Skeletonisation and leaf-wood separation}
\label{sec:skeleton}

To illustrate how the output of our pipeline can be useful for individualised ecological monitoring, we processed the output NeRF point clouds with some publicly available tools to extract the branching structure. This allows us to quantitively measure attributes of very small saplings to enable ecologists to monitor sapling health.

We first extracted the skeleton of a sapling from its point cloud $\mathbf{S}_{ij}$ using PC-Skeletor~\cite{meyer2023cherrypicker}. PC-Skeletor was originally developed to extract branch structure from tree point clouds by extracting curve-like structures from unorganised 3D data. The algorithm builds a k-nearest neighbour graph from the raw sapling point clouds and applies an iterative contraction process that progressively reduces redundant geometry while preserving global connectivity. During this process, spurious branches are pruned based on geometric saliency, resulting in a compact and topologically consistent structure.

This approach is particularly well-suited for data captured in natural forests, where point clouds are often noisy, incomplete, or occluded. Using this approach we can obtain a skeletonised point cloud $\mathbf{K}_{ij}$ that captures the sapling’s main branches, along with an open graph representation $\mathcal{G}_{ij} = (\mathbf{V}, \mathbf{E})$ encoding its topology. In Figures \ref{fig:leaf_trunk} and \ref{fig:leaf_trunk_2}, we show results of this process. It is worth noting that the point clouds obtained from NeRF are very dense. To achieve best performance with PC-Skeletor, we subsample the clouds to avoid overskeletonisation.

To separate leaf points from woody components, we initially evaluated several pre-trained models for leaf segmentation~\cite{li2023automatic,jiang2023lwsnet}. However, these models are intended for large adult trees. The methods tended to misclassify the points of a sapling point cloud $\mathbf{S}_{ij}$ and did not generalise to small-scale structures.

\begin{figure}[h]
\centering
\includegraphics[]{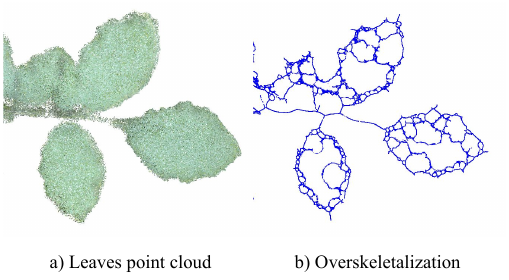}
\caption{\textbf{Leaf segmentation}: (a) A detailed view of the NeRF-derived point cloud of sapling 01 showing individual leaves. (b) Leaf segmentation with overskeletonization for segmentation purposes, which can be avoided with appropriate parameter turning. }
\label{fig:leaves}
\end{figure}

To overcome this limitation, we implemented a geometry-driven segmentation pipeline tailored to the fine-scale architecture of saplings. First, we compute the skeleton $\mathbf{K}_{ij}$ of the sapling from the high-density point cloud $\mathbf{S}_{ij}$, without applying any downsampling. This will oversegment the skeleton, as depicted in Figure~\ref{fig:leaves}. The resulting skeleton captures the main trunk and branch architecture, but due to the high surface complexity of the leaves, it also introduces a large number of terminal bifurcations at the canopy surface (Figure~\ref{fig:leaves}). These bifurcations manifest as bottom-level vertices in the skeleton graph $\mathcal{G}_{ij}$.

We then identify the corresponding leaf regions by segmenting all points in $\mathbf{S}_{ij}$ that lie within a spatial neighbourhood around these terminal vertices. This results in two disjoint point clouds: $\mathbf{L}_{ij}$ for the leaf points, and $\mathbf{W}_{ij}$ for the wood components (trunk and branches). The result of this segmentation is shown in Figures \ref{fig:leaf_trunk} and \ref{fig:leaf_trunk_2}.

This approach allows our method to extract the structural attributes that are critical for understanding ecological processes. Specifically, we derive stem height from the original point cloud $\mathbf{S}_{ij}$, branching architecture from the skeleton $\mathbf{K}_{ij}$, spatial leaf distribution from the segmented leaf points $\mathbf{L}_{ij}$, and leaf-to-wood ratio by comparing $\mathbf{L}_{ij}$ and $\mathbf{W}_{ij}$.

\begin{figure*}[ht]
\centering
\includegraphics[]{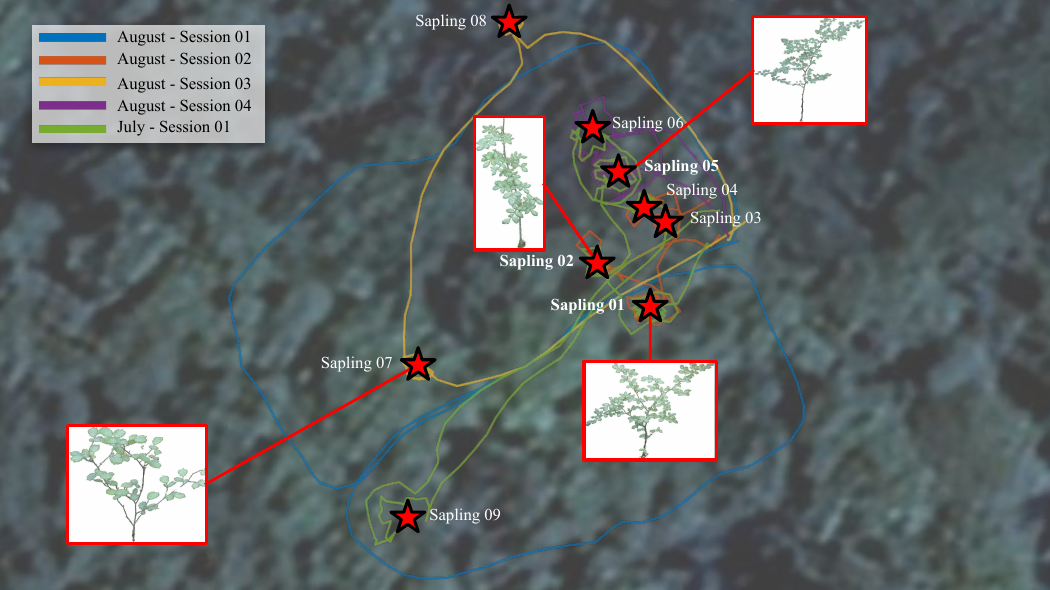}
\caption{\textbf{Multi-session geo-localisation results}: Geo-referenced aerial image of the area in Wytham Woods containing saplings 01–09. The saplings 10-13 are in another plot, 100 meters further west. The trajectories of each session are shown in different colours, combining data from July and August 2025. Each sapling is georeferenced and marked with a red star. Example point clouds generated using NeRF for several geo-localised saplings are also displayed.}
\label{fig:geo_loc}
\end{figure*}

\section{EVALUATION}
\label{sec:evaluation}

The experimental evaluation was conducted using our self-developed Frontier device, which integrates a Hesai Pandar QT64 LiDAR with an IMU, and three RGB cameras from E-CON (with model number e-CAM20\_CUOAGX). Data collection took place in Wytham Woods (UK) during July, August and December 2025 (so covering Summer and Winter) and recorded a total of 23 saplings. Five individual saplings were captured in the three recordings to demonstrate reacquisition of individual sampling for temporal and inter-seasonal monitoring. In addition, we evaluate our approach on another dataset collected in June in the Evo forest (Finland), which includes 3 conifer saplings so as to demonstrate that the approach can be used in other environment. To compare the results with the established methodology used for ecological monitoring, we scanned the saplings using a Leica RTC360 TLS scanner.

In the following, we present the results of our experimental evaluation. In Section~\ref{sec:ev_multi}, we show the evaluation of the multi-session SLAM and geo-localisation module. Section~\ref{sec:ev_novel} reports on the novel-view synthesis performance, highlighting the quality of the rendered views. Section~\ref{sec:ev_pc} focuses on the reconstruction of 3D point clouds and their applicability for long-term monitoring, assessing both geometric quality and temporal consistency. Finally, Section~\ref{sec:ev_attributes} examines the extraction of sapling architectural attributes relevant to ecological monitoring, including stem height, branching patterns, and leaf-to-wood ratios.

\subsection{Evaluating multi-session SLAM and geo-localisation}
\label{sec:ev_multi}
To assess the multi-session SLAM module, multiple mapping sessions were fused using the method described in Section \ref{sec:slam}. They were also geo-referenced using GNSS measurements. Consequently, each NeRF-derived sapling reconstruction can then be accurately positioned in a common global reference frame. Multi-session geo-localisation results are presented in Figure~\ref{fig:geo_loc}, showing the locations of saplings 01–09 overlaid on a geo-referenced aerial image of the forest area at Wytham Woods (saplings 10-13 are 100 meters west). The trajectories of each session, collected in July, August and December 2025, are depicted with distinct colors, while each sapling is marked with a red star. Example NeRF-derived point clouds for several geo-localised saplings are also shown. These results demonstrate how the proposed framework can support long-term monitoring and larger scale deployment in future.

\begin{figure}[t]
\centering
\includegraphics[]{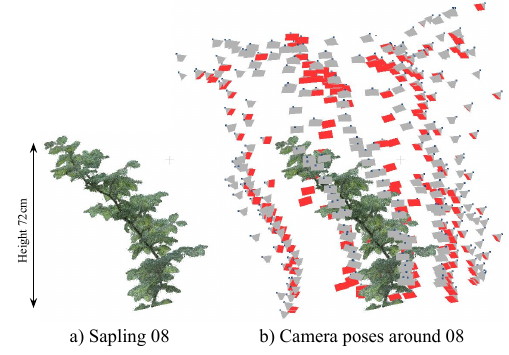}
\caption{\textbf{Camera trajectory for image acquisition}: (a) Example 3D reconstruction (of sapling 08). (b) Camera poses around the sapling, where blue points indicate the camera focal point and red/grey rectangles represent the camera frustrums.}
\label{fig:camera-poses}
\end{figure}

\begin{figure*}[t]
\centering
\includegraphics[]{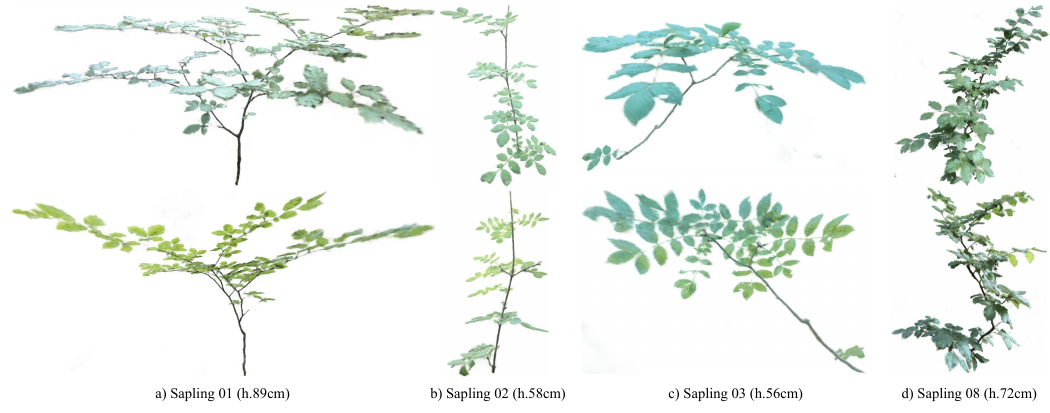}
\caption{\textbf{Novel-view synthesis results for four saplings from Wytham Woods}. Each sapling is shown from two perspectives: the top row shows views from above while the lower rows shows views from the side. Background pixels have been removed using a bounding box when rendering each tree to better highlight the reconstructed structure.}
\label{fig:novel_views}
\end{figure*}

\subsection{Evaluating novel-view synthesis accuracy}
\label{sec:ev_novel}

In Figure \ref{fig:camera-poses}, we show an example of a trajectory recorded around a sapling using the handheld device. These camera poses are used to train the NeRF model of the saplings, as shown in Figure~\ref{fig:novel_views}. In Table~\ref{tab:novel_view_eval}, we summarise the quantitative evaluation metrics (PSNR, SSIM, and LPIPS) obtained for each sapling. The saplings are clustered into three size ranges. We can see that our method performs best for smaller saplings $<$\SI{1}{\meter}. This is because it is easier to record a full set of views around the saplings for a smaller sapling (as in Figure \ref{fig:camera-poses}). Additionally, since the sequences were recorded in an object-centric manner and metrics where obtained from images including background, part of the variability in performance may be attributed to differences in the background composition rather than the saplings themselves. This effect can be observed in Figure~\ref{fig:novel_views}, where Sapling 03, which achieved the highest PSNR, appears qualitatively comparable to Sapling 08, whose reconstruction appears to be sharper despite lower PSNR.

These results confirm the robustness of our approach for novel-view synthesis, showing that despite background complexity, the reconstructed saplings maintain a consistent level of visual fidelity suitable for downstream ecological analysis.

\begin{table}[ht]
	\caption{Evaluation of Novel View Synthesis Quality}
	\centering
	\begin{tabular}{ c c c c c}
		\toprule
		Size Range (m) & Sapling ID & PSNR$\uparrow$ & SSIM$\uparrow$ &  LPIPS$\downarrow$ \\
            \midrule
		& 02 & 19.72 & 0.571 & 0.369  \\
		& 03 & 21.12 & 0.737 & 0.296   \\
	0.5 - 0.75	& 04 & 19.57 & 0.638 & 0.367   \\
		& 08 & 15.36 & 0.360 & 0.603   \\
		& 12 & 19.87 & 0.518 & 0.395   \\
        \midrule
            & 01 & 19.18 &0.643 &0.384   \\
            & 07 & 15.49 &0.429 &0.639   \\
	0.75 - 1.0	& 09 & 19.21 & 0.589 & 0.324   \\
		& 10 & 18.44 & 0.541 & 0.378   \\
		& 11 & 20.31 & 0.592 & 0.312   \\
        \midrule
        & 05 & 17.88 & 0.456 & 0.466   \\
	$>$1.0	& 06 & 17.23 & 0.412 & 0.487  \\
		& 15 & 16.24 & 0.427 & 0.526   \\
        \bottomrule
            \addlinespace
    \end{tabular}
	\label{tab:novel_view_eval}
\end{table}


\begin{figure}[ht]
\centering
\includegraphics[width=0.5\textwidth]{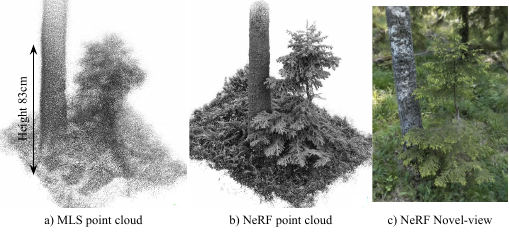}
\caption{\textbf{Comparison between reconstructions of a conifer sapling from the Evo forest (Finland)}: a) MLS point cloud, B) NeRF-derived point cloud, and c) NeRF-derived novel-view rendering. The MLS result is very noisy and is unable to represent the details of the sapling structure, while the NeRF provides a much clearer and more coherent reconstruction.}
\label{fig:lidar_vs_nerf}
\end{figure}

\begin{figure*}[t]
\centering
\includegraphics[]{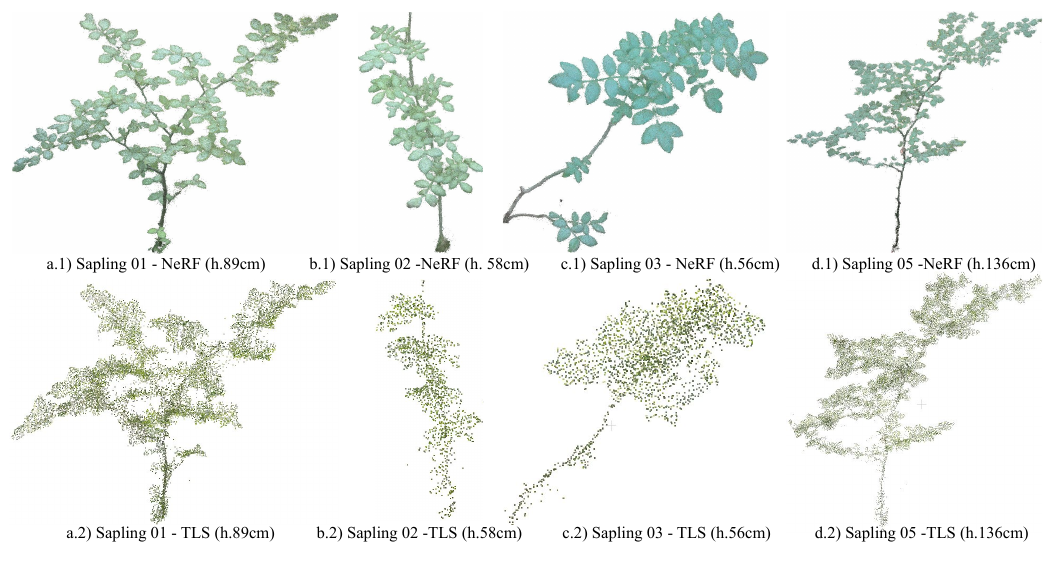}
\caption{\textbf{Comparison of NeRF-derived (top row) and TLS (bottom row) point clouds for four saplings (01, 02, 03, and 05) from Wytham Woods.} NeRF-derived reconstructions capture fine-scale geometry and foliage much more accurately than TLS, with the superiority being especially evident for smaller saplings (e.g., Sapling 03 and 02), where TLS fails to recover thin branches and leaves.}
\label{fig:point_clouds}
\end{figure*}

\subsection{Point cloud accuracy and usefulness for temporal monitoring}
\label{sec:ev_pc}

From the reconstructed NeRFs, we generate 3D point clouds that can then be used for structural and temporal analysis. Figure~\ref{fig:lidar_vs_nerf} presents a representative example of Sapling 14 from the Evo forest (Finland), comparing the MLS reconstruction, the NeRF-derived point cloud, and a novel-view rendering. In this case, our MLS pipeline is clearly too noisy and is unable to represent the fine structure of the sapling\footnote{It should be noted that other LiDAR sensors (such as Hesai's XT32 and Velodyne) are more accurate sensors than the Hesai QT64 used here.}, whereas NeRF produces a much more detailed reconstruction, capturing both stem and branch geometry as well as realistic appearance in the rendered views.

Beyond individual examples, we systematically evaluate the quality of NeRF-derived point clouds against the Leica TLS scans (which is much more accurate than the LiDAR sensor in the MLS device). Figure~\ref{fig:point_clouds} presents views for four saplings (01, 02, 03, and 05), demonstrating that NeRF-derived reconstructions are consistently superior in detail and fidelity, particularly for smaller saplings, while the TLS results are sparse and noisy. 

Sapling 03, which measures only \SI{56}{\centi\meter} in height, is much more poorly reconstructed by the TLS, while the NeRF preserves both stem and branching structures. For taller saplings such as Sapling 05 ($>$\SI{1}{\meter}), TLS can capture the structure better and some branches can be distinguished, but NeRF still resolves fine structure more clearly.

\begin{figure}[ht]
\centering
\includegraphics[height=9cm]{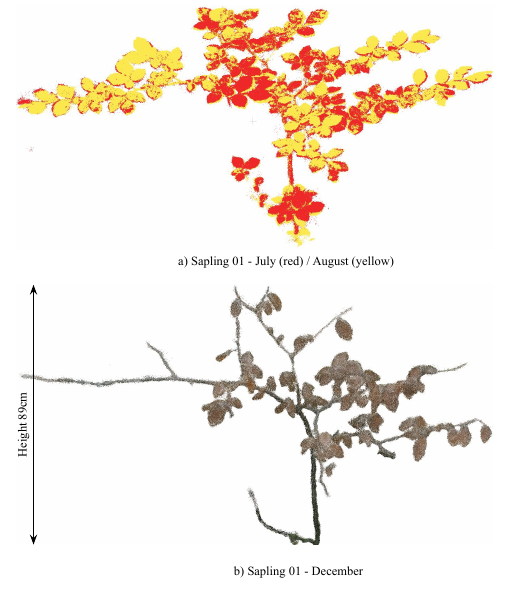}
\caption{ Upper: (a) Multi-session monitoring of Sapling 01 using NeRF point clouds for July (red) and August (yellow) - showing little change. Lower: (b) Here Sapling 1 exhibits more differences with part of the leaves fallen.}
\label{fig:longitudinal}
\end{figure}

To assess the potential of using NeRF-derived point clouds for long-term monitoring, we analysed data recorded over three recordings - from summer until winter. Figure~\ref{fig:longitudinal} (a) shows the Sapling 01, reconstructed in July (red) and August (yellow), and overlaid for comparison. Only small variations in leaf distribution are observed between the two captures. Moving to December in Figure~\ref{fig:longitudinal} (b), we see the representation of Sapling 01 where the leaves have mostly fallen.

\begin{figure}[ht]
\centering
\includegraphics[height=5.5cm]{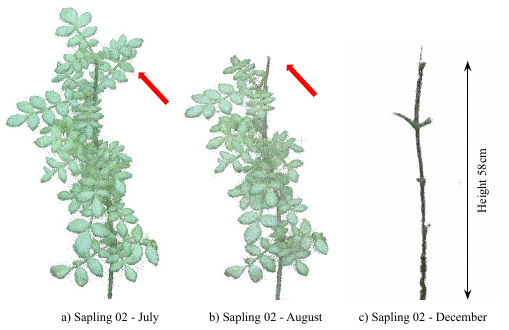}
\caption{Sapling 02 undergoing structure damage and leave off: the upper stem and highest branch was damaged between July and August (red arrows). By December the leaves have fallen off.}
\label{fig:longitudinal2}
\end{figure}

In Figure~\ref{fig:longitudinal2} we see Sapling 02 suffering from more significant branch damage. Moving from July to August a single branch (red arrow) is missing however by December, all the leaves have fallen off. This demonstrates the ability of our approach to track sapling growth but abrupt structural damage relevant to ecological monitoring.

These experiments clearly demonstrate that MLS is unable to measure elements smaller than about 3cm. NeRF consistently delivers denser, cleaner, and more faithful reconstructions of small saplings, enabling the recovery of architectural details that are otherwise lost in TLS point clouds. Even for taller saplings, where TLS performs better, NeRF still provides a noticeable improvement in geometric accuracy and completeness. This demonstrates the potential of using NeRF-derived reconstructions as a scalable and reliable alternative to traditional TLS methods for ecological monitoring of small trees.

\begin{figure*}[t]
\centering
\includegraphics[]{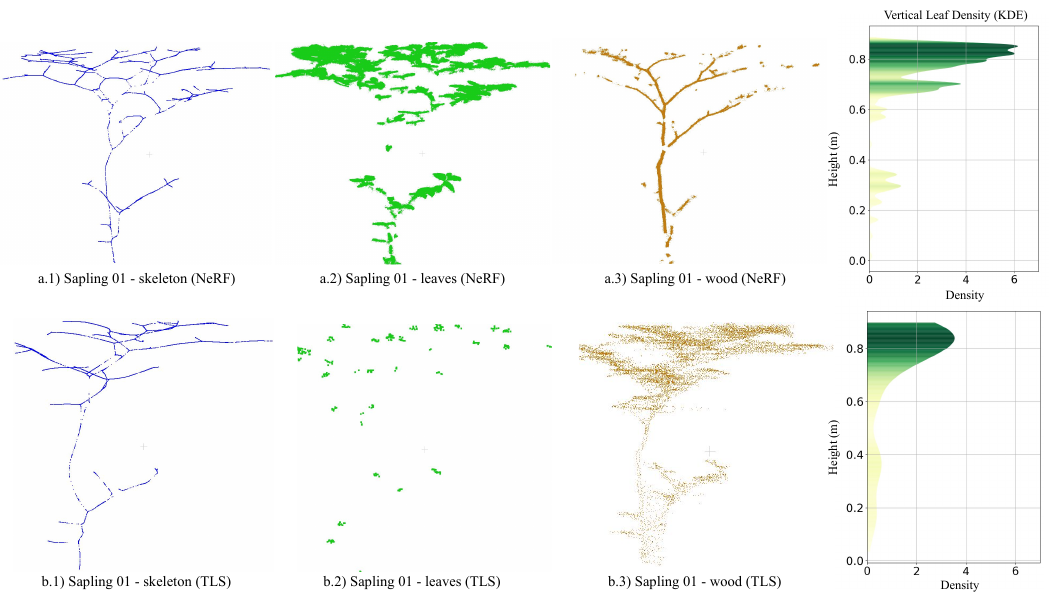}
\caption{\textbf{Attributes of Sapling 01 - using either a NeRF (top row) or a TLS pipeline (bottom row).} The subfigures show the skeleton, leaf points, wood points (left to right) as well as a vertical leaf distribution plot. TLS points are misclassified as wood. Using NeRF-derived representation produces a clearer separation of leaf and wood points and a more realistic vertical distribution of foliage.}
\label{fig:leaf_trunk}
\end{figure*}

\begin{figure*}[t]
\centering
\includegraphics[]{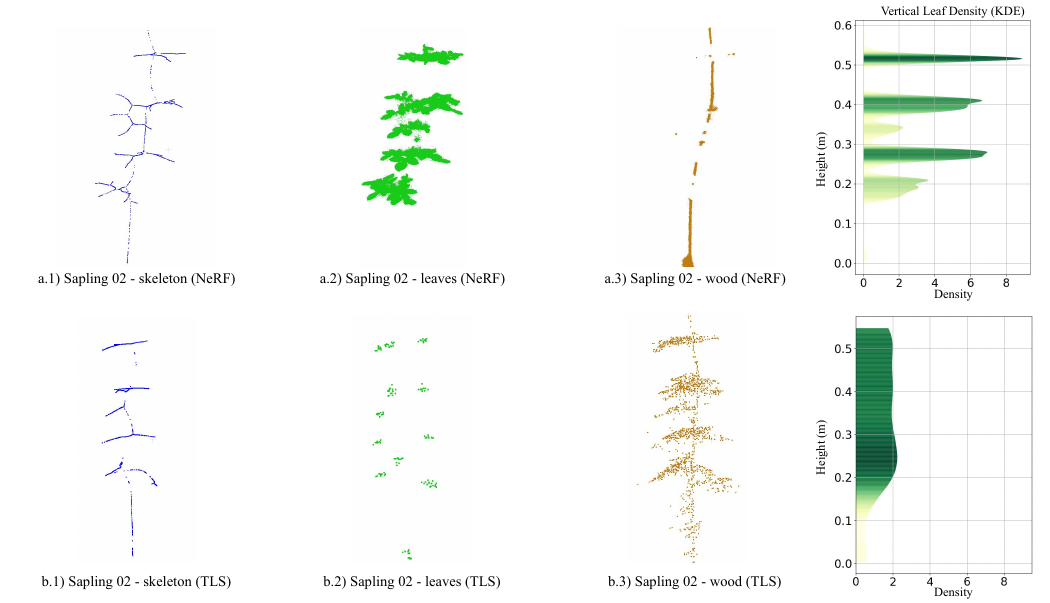}
\caption{\textbf{Measured attributes of Sapling 02 - using either a NeRF pipeline (top row) or a TLS pipeline (bottom row).} As in Figure~\ref{fig:leaf_trunk}, NeRF-derived yields a well-defined skeleton and denser leaf distribution profiles using our described procedure, whereas from TLS data, the skeletonisation pipeline has failed to extract small branches, thus classifying leaves as wood, which leads to less informative and oversimplified distributions.}
\label{fig:leaf_trunk_2}
\end{figure*}

\begin{table}[h]
	\centering
    \caption{Evaluation of sapling architectural traits}
	\begin{tabular}{|c|c c|c c|c c|}
            \cmidrule{2-7}
            \multicolumn{1}{c|}{}  & \multicolumn{2}{c|}{Height (m)} & \multicolumn{2}{c|}{Leaf Wood Ratio} & \multicolumn{2}{c|}{Bifurcations}  \\
            \midrule
		Sapling ID & NeRF & TLS &  NeRF & TLS &  NeRF & TLS \\
            \midrule
		01 & 0.89 & 0.91 & 12.54 & 0.03 & 98 & 38  \\
		02 & 0.58 & 0.59 & 23.62 & 0.09 & 27 & 11  \\
		03 & 0.56 & 0.55 & 19.24 & 0.11 & 35 & 18  \\
		04 & 0.70 & 0.72 & 17.66 & 0.10 & 47 & 22 \\
	    05 & 1.36 & 1.33 & 14.21 & 0.06 & 112 & 39  \\
	    06 & 2.89 & 2.93 & 21.24 & 0.04 & 124 & 69  \\
		07 & 0.77 & 0.76 & 18.55 & 0.12 & 84 & 35  \\
		08 & 0.73 & 0.74 & 32.59 & 0.27 & 87 & 43  \\
        09 & 0.77 & 0.79 & 21.22 & 0.15 & 78 & 37  \\
		10 & 0.85 & 0.86 & 19.87 & 0.22 & 89 & 38  \\
		11 & 0.93 & 0.91 & 22.54 & 0.13 & 101 & 46  \\
		12 & 0.61 & 0.62 & 24.18 & 0.18 & 91 & 33  \\
        \bottomrule
            \addlinespace
    \end{tabular}
    \label{tab:attributes}
\end{table}

\subsection{Sapling attributes for ecological monitoring}
\label{sec:ev_attributes}

In addition to visual and geometric evaluation, we assess the suitability of using NeRF-derived reconstructions to measure ecologically-relevant sapling attributes. Table~\ref{tab:attributes} reports a comparison between three key metrics from NeRF-derived and TLS representations: stem height, bifurcations in the sapling structure (skeleton) and the leaf-to-wood ratio (LWR), defined at the point level~as:

\begin{equation}
    \text{LWR} = \frac{N_l}{N_w},
\end{equation}

where $N_l$ and $N_w$ correspond to the number of points classified as leaf and wood, respectively. Ground truth for sapling height is unavailable, as it cannot be measured precisely with a physical ruler: the irregularity of the forest floor and the non-vertical orientation of sapling stems would lead to inconsistencies depending on the chosen reference point. Nevertheless, the estimations obtained from NeRF and TLS differ by only \SI{1}{-}\SI{2}{\centi\meter}, which indicates consistency across the methods. A larger deviation is observed for Sapling~06 ($\approx$~3~m tall), with a difference of \SI{4}{\centi\meter}. This was due to incomplete capture of the upper canopy during handheld data collection.

For the leaf-to-wood ratio, results differ significantly. The TLS-estimated value is consistently much lower, as most points are incorrectly classified as wood due to its poorly defined skeleton and limited branching detail. By contrast, NeRF-derived LWR estimates are more realistic\footnote{It is worth noting that these LWR values are indicative and not fully evaluated.}.

Additionally, for each sapling, we counted the number of bifurcations in the branch topology that describe the sapling's skeleton structure. In the case of the skeletons obtained from TLS, some fine branches are not reconstructed due to limited structural detail. As illustrated in Figure~\ref{fig:leaf_trunk}, a thin branch near the base of the sapling is fused with the trunk in the TLS skeleton, whereas it is correctly distinguished in the NeRF-derived skeleton. This effect is reflected in the increase number of detected bifurcations showin in Table~\ref{tab:attributes}.

To compare the TLS and NeRF-derived outputs, Figures~\ref{fig:leaf_trunk} and~\ref{fig:leaf_trunk_2} show examples for Saplings~01 and~02. The skeleton, leaf points, and wood points are visualised alongside the vertical leaf distribution profile. The distributions are computed using Kernel Density Estimation (KDE)~\cite{luotamo2022density,shi2025spatiotemporal}, which provides a smooth characterisation of leaf density along the stem. For both saplings, we see that the majority of the TLS points are classified as wood, resulting in overly sparse and uniform leaf distributions. For the NeRF achieves a segmentation which is denser and captures both foliage clusters and their vertical variability.

In summary, NeRF-derived reconstructions not only outperform TLS in terms of geometric fidelity but also enable the consistent estimation of ecologically meaningful attributes such as sapling height and leaf-to-wood ratio. While TLS often fails to capture fine-scale foliage and branch structures, leading to unrealistic attribute values, NeRF produces more accurate and temporally stable measurements. This capability highlights the potential of NeRF as a powerful tool for long-term ecological monitoring.

\section{CONCLUSIONS}
\label{sec:conclusions}

Saplings are fundamental indicators of forest regeneration, yet their fine-scale architectural traits are difficult to capture with conventional 3D sensing technologies. TLS, MLS, and photogrammetry are limited in their ability to reconstruct thin branches and dense foliage, and often lack the geometric consistency required for long-term monitoring. In this work, we introduced a system that fuses NeRF, LiDAR SLAM, and GNSS, to provide repeatable, geo-localised reconstructions that enable quantitative evaluation of the structure and growth of small saplings over time while localizing them within hectare-scale plots.

Our results demonstrate that the proposed approach consistently outperforms TLS and MLS for a variety of measurement tasks. In novel-view synthesis, NeRF delivers visually faithful reconstructions despite background variability. For 3D point clouds, NeRF captures greater geometric detail, especially for small saplings where TLS reconstructions often lose structural details, while also enabling long-term comparisons that reveal subtle growth patterns as well as abrupt structural changes such as branch loss. In terms of ecological attributes, we show that NeRF-derived reconstructions provide consistent height measurements (within \SI{1}{-}\SI{2}{\centi\meter} of TLS-estimated result) and more accurate leaf-to-wood ratios, thanks to the improved classification of foliage and skeletal structures. These advantages make it possible to capture ecologically relevant metrics such as vertical leaf distribution profiles with a level of detail unattainable by TLS or MLS.

Overall, the fusion of NeRF with LiDAR SLAM and GNSS represents a scalable and reliable solution for monitoring young trees \emph{in situ}. By enabling accurate reconstruction of saplings between \SI{0.5} and \SI{2}{\meter} tall, our method provides ecologists with richer structural data on stem height, branching patterns, and leaf-to-wood allocation. This capability opens new opportunities for long-term, repeatable ecological monitoring, offering quantitative insights into regeneration dynamics that are crucial for understanding forest composition, competition, and resilience.

Future work could explicitly couple the extracted sapling traits with demographic and survival models to quantify trait-mediated recruitment dynamics. Integrating species identity and functional strategies would further enable scaling the system to regeneration cohorts across disturbance gradients, positioning this framework as a foundation for trait-informed forest regeneration modelling.


\bibliography{references.bib}{}
\bibliographystyle{IEEEtran}

\end{document}